\documentclass[Afour,sage,times]{sagej}

\usepackage{moreverb,url}
\usepackage{algorithm,algorithmic}%
\usepackage{multirow}
\usepackage{xcolor}
\usepackage[normalem]{ulem}

\usepackage{amsmath}
\usepackage{array}
\usepackage[caption=false,font=normalsize,labelfont=sf,textfont=sf]{subfig}
\usepackage{float}
\usepackage{url}

\usepackage[colorlinks,bookmarksopen,bookmarksnumbered,citecolor=red,urlcolor=red]{hyperref}

\newcommand\BibTeX{{\rmfamily B\kern-.05em \textsc{i\kern-.025em b}\kern-.08em
T\kern-.1667em\lower.7ex\hbox{E}\kern-.125emX}}

\def\volumeyear{2020}

\begin{document}

\runninghead{Karim et al.}

\title{A semi-supervised self-training method to develop assistive intelligence for segmenting multiclass bridge elements from inspection videos}

\author{Muhammad Monjurul Karim\affilnum{1}, Ruwen Qin\affilnum{1}, Zhaozheng Yin\affilnum{2} and Genda Chen\affilnum{3} }

\affiliation{\affilnum{1}Department of Civil Engineering, Stony Brook University, Stony Brook, NY 11794, USA\\
\affilnum{2}AI Institute, Department of Computer Science, Department of Biomedical Informatics, Stony Brook University, Stony Brook, NY 11794, USA\\
\affilnum{3}Department of Civil, Environmental and Architectural Engineering, Missouri University of Science and Technology, Rolla, MO 65409}

\corrauth{Ruwen Qin, Department of Civil Engineering,
Stony Brook University, Stony Brook, NY 11794, USA.}
\email{ruwen.qin@stonybrook.edu}

\begin{abstract}
Bridge inspection is an important step in preserving and rehabilitating transportation infrastructure for extending their service lives. The advancement of mobile robotic technology allows the rapid collection of a large amount of inspection video data. However, the data are mainly images of complex scenes, wherein a bridge of various structural elements mix with a cluttered background. Assisting bridge inspectors in extracting structural elements of bridges from the big complex video data, and sorting them out by classes, will prepare inspectors for the element-wise inspection to determine the condition of bridges. This paper is motivated to develop an assistive intelligence model for segmenting multiclass bridge elements from inspection videos captured by an aerial inspection platform. With a small initial training dataset labeled by inspectors, a Mask Region-based Convolutional Neural Network (Mask R-CNN) pre-trained on a large public dataset was transferred to the new task of multiclass bridge element segmentation. Besides, the temporal coherence analysis attempts to recover false negatives and identify the weakness that the neural network can learn to improve. Furthermore, a semi-supervised self-training (S$^3$T) method was developed to engage experienced inspectors in refining the network iteratively. Quantitative and qualitative results from evaluating the developed deep neural network demonstrate that the proposed method can utilize a small amount of time and guidance from experienced inspectors (3.58 hours for labeling 66 images) to build the network of excellent performance (91.8\% precision, 93.6\% recall, and 92.7\% f1-score). Importantly, the paper illustrates an approach to leveraging the domain knowledge and experiences of bridge professionals into computational intelligence models to efficiently adapt the models to varied bridges in the National Bridge Inventory.
\end{abstract}


\maketitle

\section{Introduction}
The U.S National Bridge Inventory has over 600,000 highway bridges. 39\% of these bridges are over 50 years old, and almost 9\% are structurally deficient and require significant repair \cite{asce20172017}. Rehabilitation, maintenance, and rebuilding efforts are necessary for preserving the transportation infrastructure throughout the United States. For example, National Bridge Inspection Standards require that each bridge should be inspected every two years to ensure no cracks, rust, or other damages \cite{us_department_of_transportation_2017}. The conventional bridge inspection requires a crew of inspectors, heavy equipment with a lifting capability, access to dangerous heights, and the closure of the road during the time of inspection. These make the bridge inspection one of the most dangerous and costly operations in the state Departments of Transportation. Results of the visual inspection are inaccurate and vary largely among different inspectors although the image-reference approach is developed to guide the inspection \cite{MBEI2019}.

Research has taken place to develop safer and more efficient bridge inspection methods. Some adopted a completely manual approach for the bridge routine inspection, which requires a large number of inspector hours \cite{jauregui2003implementation} and inspection results vary largely among inspectors. To make inspection faster, cheaper, safer, more objective, and less interruptive to traffic, methods to automate the bridge inspection have been developed. Recently, mobile robots such as Unmanned Aerial Vehicles (UAVs) have been proven to be very helpful in dangerous, dull, or dirty applications \cite{weatherington2002unmanned}. Collecting inspection video data using aerial platforms reduces or eliminates the labor-intensive onsite inspection process and allows inspectors to assess bridges from a safer location. Yet, the use of robotic inspection platforms has solved just part of the above-discussed issues, efficient, reliable analysis of inspection video is another important task.

Letting inspectors watch the collected videos for hours and days are inefficient. It is desired that a tool can be developed to assist inspectors in extracting structural elements from the inspection videos and sorting them out by classes. Given a such tool, inspectors can concentrate on the element-wise inspection. Besides, the rating of a bridge needs to be provided by a comprehensive assessment that evaluates the impact of defects on specific elements of the bridge \cite{fhwa2004national, MBEI2019, zhu2010detection}. This requires to spatially relate detected defects with bridge elements where the defects are located. The above-mentioned approach to the bridge condition evaluation suggests that an important step of analyzing the inspection video data is to extract and index images of bridge elements. After that, defect evaluation and interpretation will take place.

Extracting structural elements from the inspection videos and sorting them out by classes is a very challenging task for practitioners. On the one hand, there could be hours of videos that need to be analyzed for every individual bridge of inspection. Watching hours of video to locate the desired regions of interest is very cumbersome work for a human. Humans are prone to fatigue. Studies have shown that the human visual inspection accuracy declines easily in dull, endlessly routine job \cite{roland1982automated,snyder1972selection}. The inspector could easily miss elements in big video data, left there without an examination. Fatigue and boredom developed from repetitively watching hours of video data induce the bias in assessing the bridge elements and evaluating the condition of the whole bridge. On the other hand, bridge inspection videos captured by aerial inspection platforms are mainly images of complex scenes, wherein a bridge of various structural elements mix with a cluttered background. Assisting inspectors in analyzing the big complex video data is greatly desired to improve their job efficiency. The development of sensing technology and deep learning methods has significantly advanced the image analysis for defect detection \cite{yeum2015vision, xu2019surface, hoskere2020madnet, hu2021structure, dong2020review, bao2019computer, deng2020imaging}. Yet, methods to create deep learning models for defect detection and classification are not directly applicable to the research problem of this paper for various reasons. For example, many of the models require to take close-up images in a nearly uniform testing background where defects are relatively large and clear to analyze. Although deep learning models for segmenting multiclass objects from images are well developed in computer vision, extracting multiclass bridge elements from inspection videos captured by aerial robotic platforms is not a completely solved problem.

A few studies have developed a strong base for infrastructure component recognition using computer vision \cite{narazaki2018automated, narazaki2020vision}. Extracting bridge structural elements from videos captured by aerial inspection platforms are facing additional challenges \cite{yeum2019automated}. These include, but not limited to, motion blur, partial or full occlusion, illumination variation, background variation, and so on. So far, some studies \cite{kang2018t, zhu2017deep, zhu2017flow, han2016seq} have reported their successful experiences, for example, utilizing the temporal information of objects in video data. But the additional computational cost is expensive. The high accuracy of deep learning models for multiclass object detection and segmentation relies on large-scale dense annotations for model training. Yet annotating a huge amount of training data for bridge inspection is not only labor-intensive but expensive as it needs the knowledge of domain experts \cite{lin2014microsoft}. To truly assist bridge inspectors in their jobs, the burden of data annotation should not be completely passed to them. The efforts that domain experts, such as inspectors, contribute to the deep learning model development must be well controlled and best utilized. The strict budget for inspector-annotated training data and the high requirement on model performance motivate the combination of self-training and active learning to create a new model training approach, which are delineated in the next section.

This paper proposes a cost-effective method to create an assistive intelligence model for detecting and segmenting multiclass structural elements from bridge inspection videos captured by an aerial inspection platform. Achieved job efficiency and the quality of the model let inspectors truly benefit from the technology advancement in their jobs. The assistive intelligence model is not an artificial intelligence model isolated from users. Instead, inspectors provide their expertise to guide the development of a deep neural network, which assures the network quickly converges to a satisfactory tool for assisting themselves in analyzing the videos of any intended bridge of inspection. Filling the gaps identified in this paper, the proposed method has anticipated technical contributions in three-fold: (i) a quick transfer of an existing deep learning network to the task of detecting and segmenting multiclass structural elements from bridge inspection videos, (ii) the use of a lightweight temporal coherence analysis to recover false negatives and identify weakness that the network can learn to improve, and (iii) the development of a semi-supervised self-training (S$^3$T) algorithm that keeps human-in-the-loop to efficiently refine the deep neural network iteratively.

The remainder of this paper is organized as the following. The related work is discussed in the next section. Then, the proposed method to create the assistive intelligence model is delineated. After that, results from evaluating the proposed method and the developed model are discussed. In the end, conclusions and future work are summarized.

\section{Literature Review}
\label{sec:RelatedWorks}

Being an important step before the detailed damage assessment, extracting regions of interest from inspection video data is receiving attention from the bridge health monitoring community. A few studies have developed a strong base for infrastructure component recognition using computer vision. For example, Narazaki et al. \cite{narazaki2018automated} used multi-scale convolutional neural networks to perform the pixel-wise classification and smoothed the segmentation result using conditional random forest. They used the scene classification result to help reduce false positives of bridge components in complex scene images. Recently, authors from the same research group \cite{narazaki2020vision} further examined two semantic segmentation algorithms and three approaches to integrating a scene classifier and a bridge component classifier. This study found that the sequential configuration outperforms other configurations if the input is complex scene images. 
Yeum et al. \cite{yeum2019automated} discussed various difficulties in analyzing inspection video data collected by aerial platforms and proposed a Convolutional Neural Network (CNN) based approach to locate and extract regions of interest from images before performing the damage detection. The study demonstrated the implementation of the developed neural network in finding candidate image patches of welded joints of the truss structure. It also showed that detecting highly relevant structural elements can greatly reduce the false positive and false negative detection in the following step of damage assessment. Yet, detecting and segmenting multiclass structural elements from inspection videos collected by aerial platforms is still not solved completely.

On multiclass object detection, the Region-based CNN (R-CNN) \cite{ girshick2014rich} has shown success in many applications. The R-CNN uses the selective search \cite{uijlings2013selective} to generate region proposals to find objects in an image. The Faster R-CNN \cite{ girshick2015fast} was proposed to make the R-CNN faster. It offers improvements in both speed and accuracy over its predecessors through the shared computation and the use of a neural network to propose regions. Then, the Mask R-CNN \cite{he2017mask}, an extension of the Faster R-CNN, was proposed to perform the bounding box regression and the pixel-level segmentation simultaneously. The R-CNN and Mask R-CNN models work well in detecting objects from static images. But results may not be consistent when they process video data. Therefore, the temporal coherence information of objects in successive frames has been introduced to address the issue of inconsistent detection \cite{kang2018t,zhu2017deep, zhu2017flow}, wherein the tubelet and optical flow are used to propagate features from one frame to another. Temporal coherence analysis methods in the literature are computationally expensive due to the requirement for repeated motion estimation and feature propagation. Seq-NMS \cite{ han2016seq} has a modification only in the post-processing phase and, thus, it is faster than others. However, seq-NMS tends to increase the volume of false positives because it neither puts any penalty on false positives nor adds additional constraints to prevent the occurrence.

Creating a deep neural network usually requires a huge amount of annotated data for model training. The manual annotation of data is not only a costly process but often prone to errors. To overcome this issue, transfer learning is introduced to structural damage detection \cite{zhang2018bridge,gao2018deep,zhang2019surface,gopalakrishnan2018crack}. With transfer learning, only a relatively small dataset is needed to refine an existing deep network, which reduces the training time while keeping a good performance. Another way to tackle this annotation problem is to use semi-supervised learning that requires some labeled and some unlabeled training data. Papandreo et al. \cite{papandreou2015weakly} developed a method requiring a small number of strongly annotated images and a large number of weakly annotated images for training. They used an expectation maximization method to generate the pixel-level annotation from weakly annotated training data. Mittal et al. \cite{mittal2019semi} proposed an approach that relies on adversarial training with a feature matching loss to learn from unlabeled images. Some researchers used the self-training, a wrapper based semi-supervised method \cite{triguero2015self}, which starts training a network with only a few annotated samples and then let the network automatically annotate more training samples. This technique has been applied to a variety of image/video processing applications \cite{hung2018adversarial, misra2015watch} to reduce the effort of human annotation. While most semi-supervised learning methods save the data annotation effort compared to supervised learning, their performance is still not good enough. The primary reason for this challenge is the quality of the automatically annotated data. The inclusion of some samples mislabeled by the network itself may sharply deteriorate the training process.

Recently, a promising approach named active learning has been proposed to reduce the annotation cost for training deep neural networks \cite{settles2012active}. The essence of active learning is to train a network by actively selecting training samples from a pool of unlabeled data, which will be labeled by a human annotator to re-train the model. The fundamental assumption underlying this approach is that selecting fewer but informative data and allowing the model to learn with it may achieve greater performance than training the network using a large amount of labeled data. The performance of active learning depends on the samples selected from the unlabeled data. Traditional sampling methods do not always guarantee to provide the most representative training samples in support of active learning because of the dataset diversity and limited knowledge about the dataset. Many strategies have been developed for sampling data from unlabeled data for active learning \cite{settles2012active}. Uncertainty sampling is the most widely used strategy, which queries for samples that it is most unsure about. For example, Tian et al. \cite{tian2020joint} used clustering and the fuzzy-set selection method to choose the most uncertain and informative samples. This method increases the training sample diversity. The use of active learning in image classification has been widely explored \cite{beluch2018power, sener2018active}. However, the potential of active learning is less thoroughly explored in the more complex task of instance segmentation that usually has a relatively higher annotation cost. Morrison et al. \cite{morrison2019uncertainty} considered both spatial and semantic uncertainties of prediction using the dropout sampling. By adding dropout layers to the fully connected layers of the network, multiple times of inference over the same image are made to measure the segmentation uncertainty. A similar approach was adopted by Gal et al. \cite{gal2017deep} who used multiple forward passes with dropout at the inference (Monte Carlo dropout) to obtain better uncertainty estimates for instance segmentation. Yang et al. \cite{yang2017suggestive} trained a set of fully convolutional networks iteratively and estimated uncertainty and similarity from an ensemble of networks to determine new data for annotation. The above-discussed studies demonstrated that determining the most informative samples for active learning is complex. Besides, uncertainty-based approaches can be prone to querying outliers \cite{siddiqui2020viewal}. Simple but effective methods for recommending new data for annotation are greatly desired.

\section{Methodology}
\label{sec:Methodology}
\begin{figure}[tb]
\centering
\includegraphics[width=\linewidth]{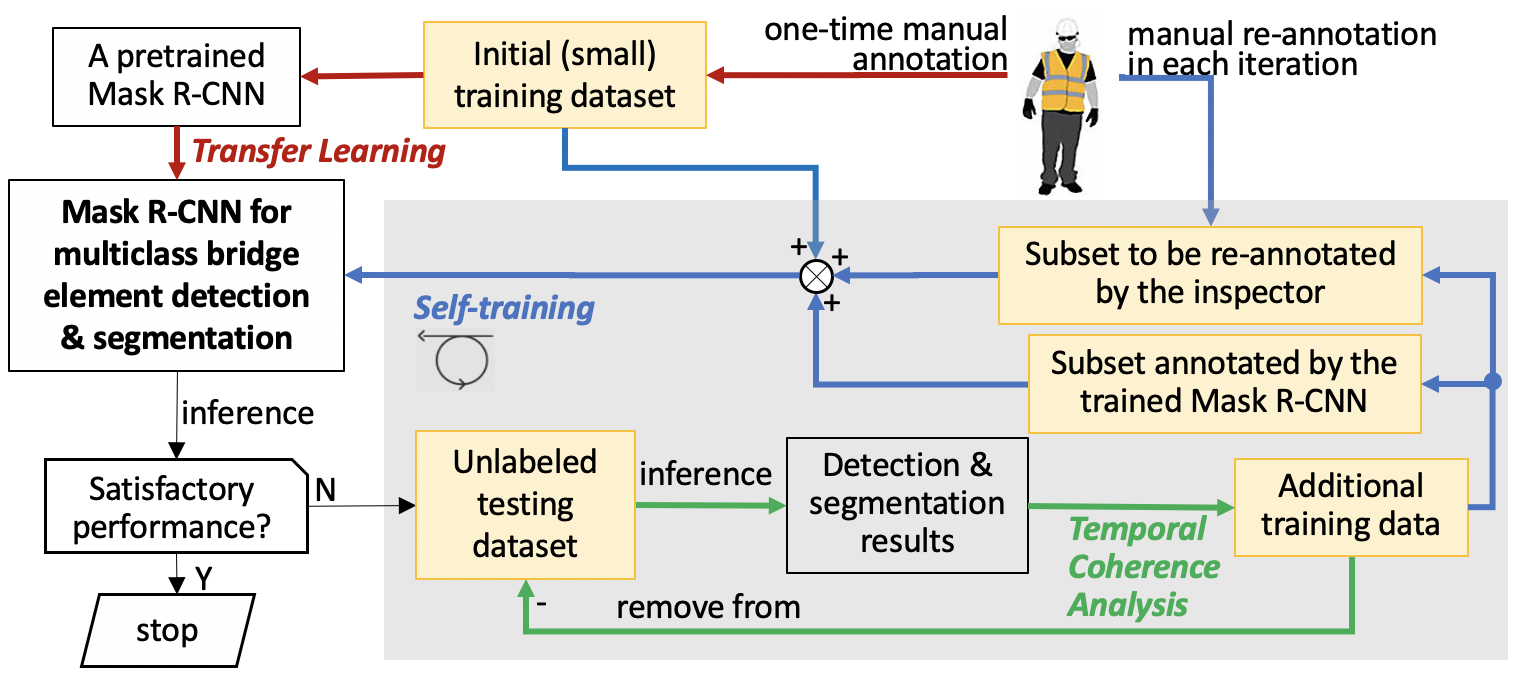}
\hfil
\caption{Overview of the proposed S$^3$T method with human-in-the-loop.} 
\label{fig_1_overview}
\end{figure}

The proposed method to create the assistive intelligence model for the multiclass bridge element segmentation is illustrated in Figure \ref{fig_1_overview}. First, a pre-trained Mask R-CNN has been chosen. Then, a small set of initial training data, which are annotated by the inspector, is used to fine-tune the network to transfer it for the new task of multiclass bridge element segmentation. The transferred network will be improved iteratively until it achieves the satisfactory performance. The shaded portion in Figure \ref{fig_1_overview} is the iterative process for performance boosting. In each iteration of the semi-supervised self-training (S$^3$T) with human-in-the-loop, the network that has not reached the satisfied performance will be applied to an unlabeled testing dataset to obtain the detection and segmentation results. Temporal coherence analysis of the results is performed to recover false negative results that are hard data for the network. After that, a small set of the recovered hard data is selected as additional training data that are removed from the unlabeled testing dataset. The additional training data is then split into two subsets, one has been automatically annotated by the network trained from the current iteration and the other subset is manually re-annotated by the inspector. The additional training data along with the initial small training dataset is then used to re-train the Mask R-CNN in the next iteration to boost its performance. The S$^3$T method let the inspector annotate selected samples that the current network failed to detect. Through learning from its weakness, the performance of the network increases quickly after a few iterations.

\subsection{Adapting the Deep Neural Network to a New Task Through Transfer Learning}
\begin{figure}[htb]
\centering
\subfloat{\includegraphics[width=\linewidth]{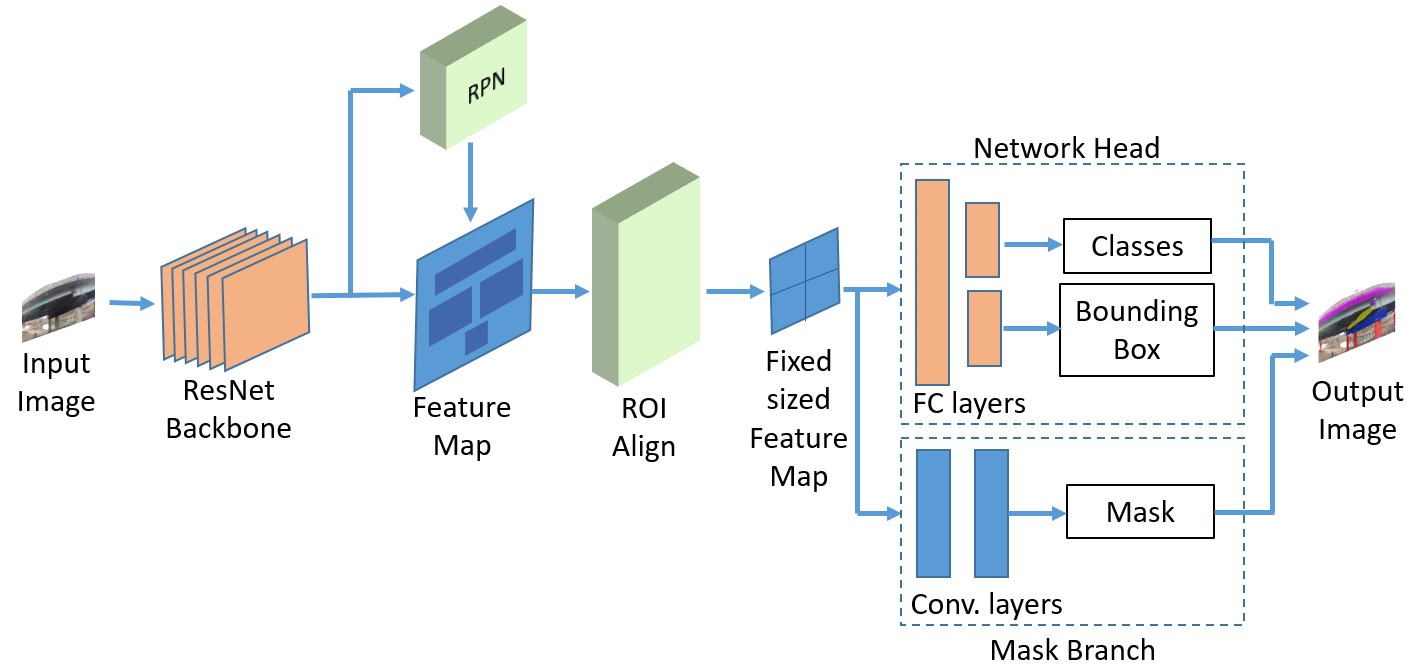}}
\hfil
\caption{The architecture of Mask R-CNN that performs the detection and segmentation of multiclass bridge elements.}
\label{fig_2_MRCNN}
\end{figure}

This study chose a Mask R-CNN as the tool for detecting and segmenting bridge elements from inspection video data. Figure \ref{fig_2_MRCNN} illustrates the structure of the Mask R-CNN. Video data are input into the network frame by frame following their order on the timeline. The backbone of the network is a feature extractor that generates the feature map of each input image. A region proposal network (RPN) creates proposal boxes named anchors and predicts the possibility of an anchor being a bridge element. Then, the RPN ranks anchors and proposes those most likely containing bridge elements, which are termed RoIs. A layer named Region of Interests Align (RoIAlign) extracts the region of interests (RoIs) from the feature map, aligns them with the input image, and converts them into fixed-size region feature maps. The fixed-size feature maps of RoIs are fed into two independent branches: the network head branch that performs the classification and bounding box generation, and the mask branch that independently generates instance masks. Readers interested in the detail of Mask R-CNN can refer to the work by He et al.\cite{he2017mask}.

Training the Mask R-CNN for the new task of multiclass bridge element segmentation from the scratch requires a large volume of annotated data to achieve a satisfying prediction accuracy.
This task does not have a large volume of annotated data for model training. To obtain high quality annotated data for this task requires the knowledge of professionals in the domain of study. Only bridge inspectors are confident in annotating bridge elements from the inspection videos. In this study, transfer learning is first used to tackle this challenge, which improves learning of the new task by transferring knowledge from a related task that has already been learned \cite{goodfellow2016deep}.

The Mask R-CNN in this study was initialized by adopting the ResNet-50 feature extractor \cite{he2016deep} whose weights have been pre-trained on the Microsoft COCO dataset consisting of more than 120,000 labeled images and around 1.5 millions of object instances in 80 categories \cite{lin2014microsoft}. Then, transfer learning was used to adapt this feature extractor to the setting of bridge inspection. Specifically, the ResNet-50 was fine-tuned using a small set of training data $(T_0)$ with a portion collected from the intended bridges of inspection. The detail of the fine-tuning process will be presented in the next section.

\subsection{Temporal Coherence Analysis for Recovering False Negative Results}
\label{subsubsection:temporal}
Mask R-CNN is a static image detector in that it processes individual images independently. When it is applied to frames of a video stream, false negative results are likely to happen due to sudden scale changes, occlusion, or motion blur. This study used the temporal coherence information of objects in successive frames to recover false negative detections and segmentations.

\begin{figure}[tb]
\centering
\includegraphics[width=\columnwidth]{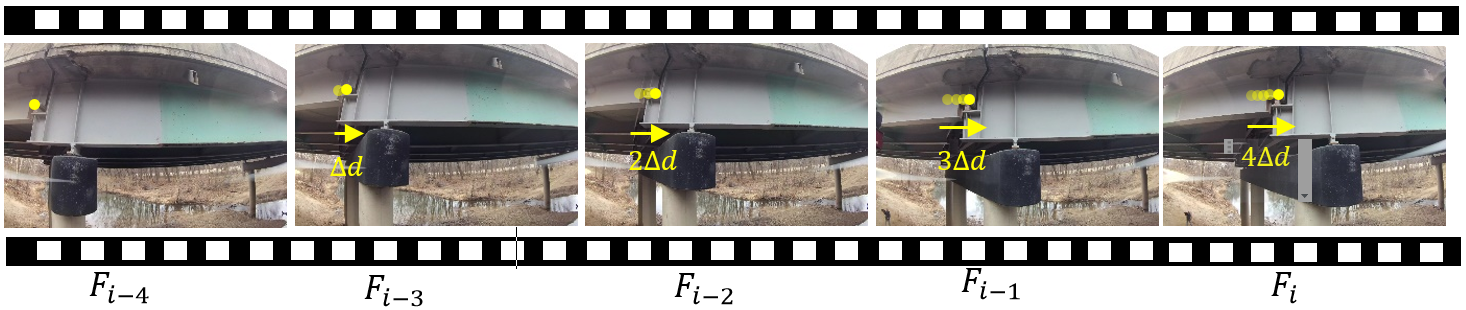}
\caption{An illustration of spatial displacements.} 
\label{fig_3_spatial_displacement}
\end{figure}

Consider a video clip that consists of a series of \textit{N} frames, indexed by $i$. In each frame the network returns ${M_{i}}$ objects with segmentations, indexed by \textit{j}. An object in a frame is highly likely to present in its neighboring frames within a range of displacement. Let, ${o_{i,j}}$ designate object $j$ in frame $i$. The center of the bounding box for ${o_{i,j}}$ is specified by its coordinates ${C_{i,j}}= ({x_{i,j}},\, {y_{i,j}})$. In $p$ frames, ${C_{i,j}}$ may shift to a surrounding pixel within a spatial displacement of ${p}\Delta d$ where $\Delta d$ is the maximum displacement between two consecutive frames. $\Delta d$ is affected by both intrinsic and extrinsic camera parameters, as the Appendix explains. $\Delta d$ is proportional to the focal length of camera and the maximum displacement of the moving camera between capturing two successive frames of image; it is inversely proportional to the size of pixels in the images taken by the camera and the distance of the camera to the object along the optical axis. The study roughly estimated $\Delta d$ according to its formula in Eq. (\ref{eq:Deltad}) using partial information and then improved the estimation experimentally by reviewing the inspection data.A value of 60 pixels was found to be appropriate in this study. Figure \ref{fig_3_spatial_displacement} illustrates an example wherein a joint of the bridge in frame $i-4$ is also shown in the succeeding four frames but with displacements.

The algorithm of temporal coherence analysis for recovering false negative results is summarized as the pseudo-code in Algorithm \ref{algorithm1} and explained below. The prediction threshold is set as a range [${t_{l}}, {t_{u}}$]. An object with a prediction score within this range is possibly a false negative prediction. Let $S_{i,j}$ denote the prediction score for object $o_{i,j}$. The network immediately returns a positive result if ${S_{i,j}} \geq {t_{u}}$ and will not predict any object if ${S_{i,j}} < {t_{l}}$. Let ${O_{i}}$ be the set of confidently predicted objects in frame $i$. The prediction score and the center location of these objects, $\{({S_{i,j}}, \, {C_{i,j}})|o_{i,j}\in O_i\}$, are the temporal coherence information for analyzing the succeeding $k$ frames. That is, $k$ is the temporal window that defines the range of preceding frames where the temporal coherence analysis searches the same objects as the weakly predicted objects (e.g., they are possibly false negative results) in the current frame. If $ {t_{l}} \leq {S_{i,j}} < {t_{u}}$, the weakly predicted object ${o_{i,j}}$ is checked by referring to a pair of preceding successive frames up to $k-1$ times, starting from the nearest pair (frames $i-1$ and $i-2$) to the farthest pair (frames $i-k+1$  and $i-k$). If an object of the same class as ${o_{i,j}}$ is found in both frames $i-1$ and $i-2$ (i.e., there exists $o_{i-1,j'}\in O_{i-1}$ and $o_{i-2,j''}\in O_{i-2}$ such that $o_{i-1,j'}=o_{i-2,j''}=o_{i,j}$), and the spatial displacements of ${o_{i,j}}$ from $o_{i-1,j'}$ and $o_{i-2,j''}$ are small, within $\Delta d$ and $2\Delta d$, respectively, this weakly predicted object is determined as a false negative prediction. The false negative prediction is recovered by adding it to ${O_{i}}$ and updating its score to be the average score of $S_{i-1,j'}$ and $S_{i-2,j''}$. Otherwise, ${o_{i,j}}$ is searched in ${O_{i-2}}$ and ${O_{i-3}}$ to determine if it is a false negative result that can be recovered. This search will continue as needed.  If ${o_{i,j}}$ is not found to be a positive result with confidence in the neighboring frames after $k-1$ times of temporal coherence analysis, it will be eliminated from the candidate list. Using a pair of frames as the reference instead of referring to a single frame will make the temporal coherence rule more strict and minimize the risk of progressively propagating a false positive result in a single frame to the target frame.

\begin{algorithm}
\caption{Temporal Coherence Analysis for Recovering False Negative Results.}
\begin{algorithmic}
\STATE // $N$: the number of video frames;
\STATE // $M_i$ : the number of objects in frame $i$;
\STATE // $O_i$ : the set of confidently predicted objects in frame $i$;
\STATE // $S_{i,j}$: the prediction score of the $j$th object in frame $i$;
\STATE // $C_{i,j}$: the center location of the $j$th object in  frame $i$;
\STATE // $t_l$ and $t_u$: prediction thresholds;
\STATE // $k$: the number of frames that have stored temporal coherence information of predicted objects.
\vspace{0.03in}
\FOR{$i=1 \: to \: N$} 
\FOR{$j=1 \: to \: M_{i}$} 
\IF{$S_{i,j}\geq t_{u}$} 
\STATE {add object ${o_{i,j}}$ to the set ${O_{i}}$ with its prediction score, ${S_{i,j}}$, and the center location, ${C_{i,j}}$ }
\ELSIF{\ $S_{i,j}\geq t_{l}$}
\FOR {\ ${q = 1,2, ... (k-1)}$} 
\IF{
$\begin{aligned}
&\exists\; o_{i-q,j'}\in O_{i-q}\; \&\; o_{i-q-1,j''}\in O_{i-q-1},\quad\\
&\ni o_{i,j}=o_{i-q,j'}=o_{i-q-1,j''}\\
&\&||{C_{i,j}- C_{i-q,j'}}||_{2} \leq q\Delta d \\
&\&||{C_{i,j}- C_{i-q-1,j''}}||_{2} \leq (q+1)\Delta d,\\
\end{aligned}$
}
\STATE let  $S_{i,j} = (S_{i-q,j'}+S_{i-q-1,j''})/2,$
\STATE {add object ${o_{i,j}}$ to the set ${O_{i}}$ with its ${C_{i,j}}$ and updated $S_{i,j}$, }
\STATE \textbf{break}
\ENDIF
\ENDFOR
\ENDIF
\STATE Eliminate the low score $(<t_u)$ object $o_{i,j}$ from the candidate list.
\ENDFOR
\ENDFOR
\end{algorithmic}
\label{algorithm1}
\end{algorithm}


The appendix has explained factors that impact the choice of $\Delta d$. The choice of $k$ is relevant to $\Delta d$. Given that $\Delta d$ is 60 pixels, an object may appear in a sequence of frames because the dimension of images is way larger than 60 pixels. The object only appears in a short video clip and then it disappears because the UAV brings the camera away from the object. If $k$ is too small, for example $k=1$, the analysis does not fully utilize the temporal coherence information of objects in successive frames. If the $k$ value is too large, false positive results will be propagated to many frames. Based on the above-discussed facts, the study experimentally examined the selection of $k$ and found a $k$ value of 4 is suitable in this study.

The range of detection thresholds for the temporal coherence analysis $[t_l, t_u]$ should be appropriately chosen. The upper boundary $t_u$ should be high enough but not extremely high to properly control both types of false results. The lower boundary $t_l$ should be below the upper boundary with a sufficient span to capture most false negative results. Selecting an extremely small lower boundary just increases the workload of temporal coherence analysis, but it has minimal impact on the result due to the control effect of the upper boundary. This study chose $[0.5, 0.9]$ as the threshold range whose appropriateness was verified on small-scale experiments.

The temporal coherence analysis identifies and picks up samples that the current network fails to predict correctly. Therefore, the proposed self-training method effectively learns from its weakness in each round of iterations. Since this temporal coherence is applied in the inference stage, it is a computationally cheap approach for evaluating and sampling new data for annotation.

\subsection{Refining the Network Through Self-Training with Human-in-the-Loop}
\label{subsubsection: SSL}

After transfer learning has initialized the Mask R-CNN for the task of bridge inspection, the network may need to be further refined, for example, by adding additional training data. If the refined network has not reached a satisfying performance, the refining process will continue. To lower the cost of data annotation and, meanwhile, maintain a good quality of the training data, the study chose the S$^3$T method that engages the inspector in continuous refinement of the network. In each iteration, a set of unlabeled data is fed to the trained deep neural network to be labeled automatically. Using the temporal coherence information of predictions, hard samples are collected from this newly created labeled data. A representative subset of the hard samples is identified and added to the training dataset to refine the network. Before this subset is added to the training dataset, a portion of it is manually re-annotated by the inspector to guide the network's learning. This process continues iteratively until the network performance reaches the target. The S$^3$T method with human-in-the-loop is further summarized as the pseudo-code in Algorithm \ref{algorithm_2}. 

\begin{algorithm}[tb]
\caption{Iteratively Fine-Tuning the Network with S$^3$T.}\label{euclid}
\begin{algorithmic}
\STATE // {$l$: index of iteration};
\STATE // $T_l$: the training dataset for iteration $l$;
\STATE // $V$: unlabeled dataset for S$^3$T;
\STATE // $R_l$: the recovered hard dataset from temporal coherence analysis;
\STATE // $S_l$: a subset of $R_l$, which is sampled based on the sampling method SP(s);
\STATE // $S_{l,\alpha}$: a fraction of $S_l$ in the size of $\alpha$ automatically annotated by the trained network;
\STATE // $S_{l,(1-\alpha)}$: a fraction of $S_l$ in the size of $1-\alpha$ to be manually annotated by the inspector;
\STATE // $M_l$:  {the data annotated by the inspector and added to the training dataset in iteration $l$}.
\vspace{0.03in}

\FOR{$l \geq 0$} 
\STATE{Fine-tune the network with $T_l$,}
\STATE{\textbf{break} if the performance meets the requirement.}
\STATE{obtain $R_l$ through the temporal coherence analysis,}
\IF{$l=0$}
\STATE{sample $S_l$ from the prediction result,}
\ELSE
\STATE{sample $S_l$ from $R_l$ using the skip sampling method,}
\ENDIF
\STATE{Split $S_l$ into two mutually exclusive parts,}
\STATE{manually annotate $S_{l,(1-\alpha)} $ to obtain $M_l$,}
\STATE{$T_{l+1} = T_l \cup M_l \cup S_{l,\alpha}$,}
\STATE{increase $\alpha $ to lower the inspector's workload in data annotation if applicable.}
\ENDFOR
\end{algorithmic}
\label{algorithm_2}
\end{algorithm}

\begin{figure}[tb]
\centering
\includegraphics[width=\columnwidth]{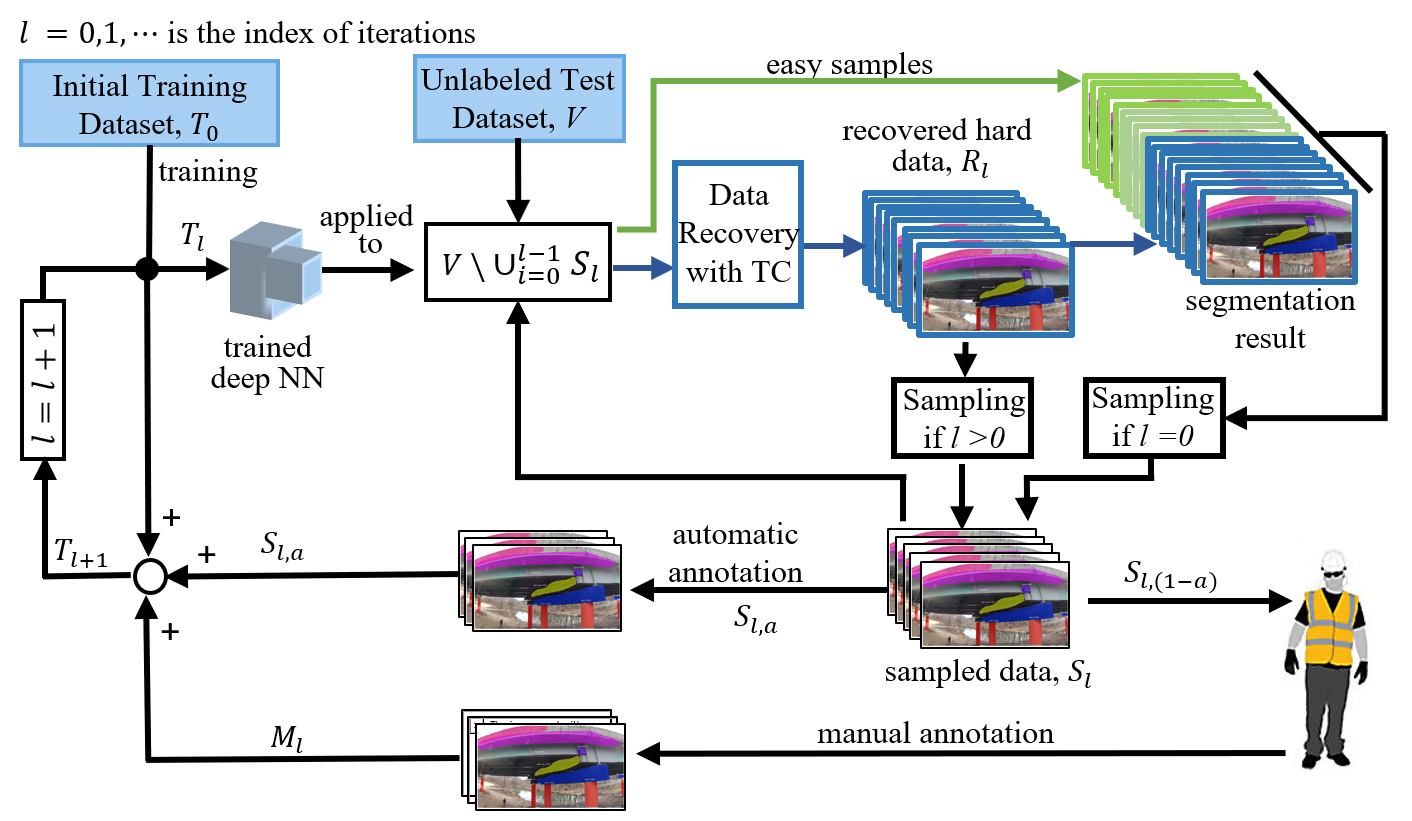}
\caption{Schematic diagram of the semi-supervised self-training (S$^3$T) for refining the Mask R-CNN model iteratively.}
\label{fig_4_iterative}
\end{figure}

Let $l$ denote the index of iterations. Figure \ref{fig_4_iterative} shows that a small training dataset $T_0$ is created to transfer the Mask R-CNN. Then, the network is applied to an unlabeled dataset, $V$. If the performance of the network is not satisfying, a portion of the prediction result is sampled as the additional training data, denoted as $S_l$. For the initial iteration when $l=0$, this $S_l$ is taken from all prediction result to guide the network. In each iteration, the selected sampled data $S_l$ is eliminated from $V$ for the further assessment of future networks. For the following iterations (i.e., $l>0$), the S$^3$T algorithm differentiates hard samples from easy samples in $V$. Easy samples are segmented by the network with relatively high reliability whereas hard samples, $R_l$, which contain a variety of situations when objects are difficult to detect, are recovered by the developed temporal coherence analysis. The sample $S_{l}$ (for $l>0$) is selected only from the recovered hard samples $R_l$. $S_l$ is divided into two mutually exclusive and collectively exhaustive subsets, $S_{l,\alpha}$ and $S_{l,1-\alpha}$, where $\alpha$ and $1-\alpha$ indicate their sizes in proportion to $S_l$. $S_{l,\alpha}$ has been automatically annotated by the trained network in testing and directly added to the training dataset. The inspector re-examines the remainder of $S_l$ and corrects false predictions, if any, before adding the inspector-annotated data $M_l$ to the training dataset. That is, at the end of the $l$\textsuperscript{th} iteration the training dataset is updated per Eq. (\ref{eq:trainingdataupdate}): 
\begin{equation}
T_{l+1} = T_l \cup M_l \cup S_{l,\alpha}.
\label{eq:trainingdataupdate}
\end{equation}

The network is re-trained using the updated training dataset and the prediction result is assessed. If the termination criterion has been met, the fine-tuning process is terminated. Otherwise, it continues refining the network. The termination criterion of the iterative process is subject to the user's choice considering the consequences of false positive and false negative results, respectively. This study chose to terminate the iterative process when both precision and recall reach 90\% or higher, and f1-score is 92\% or higher, at the Intersection over Union (IoU) value 0.5.

This iterative process has two designs: the method for sampling $S_l$ from $R_l$, and the way of determining the fraction of $S_l$ to be examined by the inspector.

\subsubsection{Skip Sampling Method, SP($s$).}
\label{subsubsec:SP}
Consecutive frames of a video are similar and, therefore, sampling a portion of frames that are evenly distributed on the timeline would be sufficient for representing the video. This study samples $S_l$ from $R_l$, for any iteration $l$, according to a skip sampling strategy $SP(s)$ that samples a frame and then skips $s$ frames. The choice of a value for $s$ needs to consider the inspection platform's speed and the camera speed. The unlabeled test dataset $V$ is a time series of $N_v$ frames. $I_{SP}$ is a $1\times N_v$ indicator vector of binary variables that define frames to be sampled according to $SP(s)$; that is, 
\begin{equation}
I_{SP}(n)=1, 
\end{equation}
for $n=1, 1+(s+1),\dots,1+(s+1)\left\lfloor N_v/(s+1) \right\rfloor$. $I_{R,l}$ is also a $1\times N_v$ indicator vector of binary variables that identify the frames recovered by the temporal  coherence analysis. The Hadamard product of $I_{R,l}$ and $I_{SP}$ yields the vector $I_{S,l}$, 
\begin{equation}
I_{S,l}=I_{R,l}\circ I_{SP},
\end{equation}
which identifies the frames to be sampled from $R_l$ according to $SP(s)$ for forming $S_l$. 

\subsubsection{Regulating the Amount of Human Guidance in S$^3$T.}
\label{subsubsec:humaneffort}
A fraction of the dataset $S_l$ from any iteration is automatically annotated by the trained network. The initial performance of the neural network is not high and data mislabeled by the network are present in $S_l$. Through examining a fraction of $S_l$ and correcting mislabeled data, the experienced inspector guides the network to quickly learn new features. $S_{l,\alpha}$ is the fraction of $S_l$ which is added to the training dataset without further human annotation. The inspector's guidance can be gradually reduced as the network starts to learn well by itself and provide improved prediction. Therefore, the fraction of automatically annotated data $S_{l,\alpha}$ can be gradually increased over iterations. Choosing the initial value of $\alpha$ is also critical as it regulates the amount of mislabeled data that may enter the training dataset when the model performance is well below the target performance. Determining an optimal selection of $\alpha$ for the S$^3$T method is a research problem but going beyond the scope of this paper. The paper illustrates the impact of choosing $\alpha$ in Table \ref{tab:alpha}.

\section{Implementation and  Result Discussion}
\label{sec:CaseStudy}
This section illustrates the implementation and evaluation of the proposed method to create the assistive intelligence model for processing the bridge inspection video data. Findings from this study are discussed.

\subsection{The Implementation Detail}
\subsubsection{The Data.}
\begin{figure}[tb]
\centering
\includegraphics[width=\linewidth]{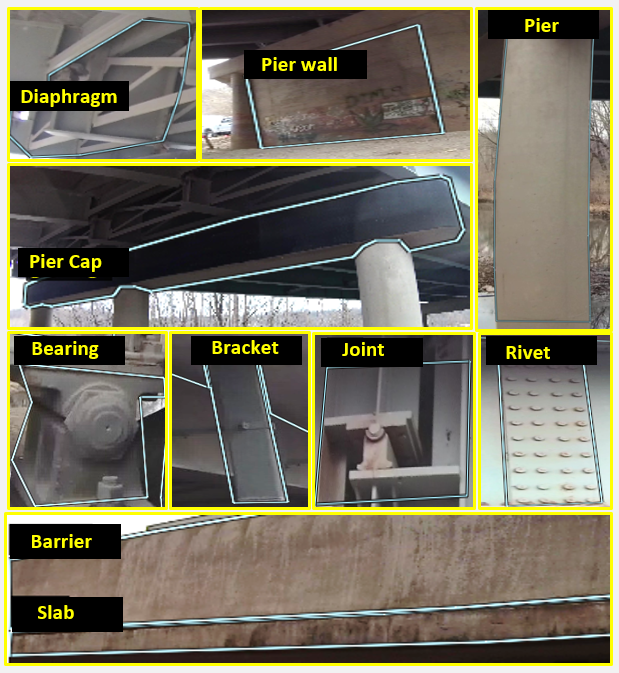}
\caption{Sample images with corresponding pixel-level object polygon with labels.}
\label{fig_class}
\end{figure}

BIRDS \cite{Chen2020}, an aerial inspection platform developed by the INSPIRE University Transportation Center, was used to capture videos of bridges in inspection. The average speed of BIRDS is 20 miles per hour (mph). The frame rate of the camera is 30 frames per second (fps) and the image resolution is 3,840 $\times$ 2,160 in pixel. A dataset $D$, which is an inspection video of 4,440 images, was used to develop and evaluate the assistive intelligence model. The initial training dataset $T_0$ contains 40 images, with 18 images from $D$ and 22 images from the inspection of other bridges. Choosing some images of other bridges adds helpful data variation to the initial training dataset. In total, the initial training dataset contains 482 objects with class labels, which are from 10 different classes of bridge elements interested to inspect. The 10 object classes are barrier, slab, pier, pier cap, diaphragm, joint, bearing, pier wall, bracket, and rivet, illustrated in Figure \ref{fig_class}. This study used the image annotation tool VGG Image annotator \cite{dutta2019vgg} to annotate labels of the objects and give pixel-level coordinates to those objects. An unannotated dataset $V$ that comprises 670 images from the dataset $D$ was particularly created for implementing the S$^3$T method. $V$ contains 5,916 objects from the 10 classes.  A test dataset $T_s$ has been created to evaluate the model performance from each training iteration. This dataset has 212 images with 1,872 objects. 

\subsubsection{Initial Adaption.}
\label{subsubsec:InitialTraining}
The proposed method was implemented by extending an existing implementation of Mask R-CNN by Matterport Inc. \cite{matterport_maskrcnn_2017}. Training and testing were performed using two Nvidia Tesla V100 GPUs with 32GB of memory. The pre-trained ResNet-50 feature extractor was fine-tuned using the initial training dataset $T_0$. Different data augmentation techniques have been applied during the training stage to improve the model's ability to generalize at various applicability where input data distortion is present. Those include horizontal flip, rotation, translation, color distortion, and random noise. The network head and the mask head (see Figure \ref{fig_2_MRCNN}) were trained for 30 epochs while keeping all the parameters of the previous layers fixed. Each epoch consists of 100 training iterations. Stochastic gradient descent was used as the optimizer and the momentum was 0.9. The learning rate of 0.001 and a batch size of 4 were used in this training process that took about 21 minutes to complete. According to Algorithm \ref{algorithm_2}, after the Mask R-CNN is transferred to have an initial adaption to the task of bridge inspection, $V$ is annotated by this network. Considering the performance of the initially adapted network, 8 images, which are about 1\%  of the images in $V$, are selected and re-annotated by the inspector and added to the initial training dataset $T_0$, becoming $T_1$, the training dataset for the next iteration. These 8 images are excluded from $V$ for further iterations. Letting the inspector to check a small amount of the prediction result of the initially adapted network in accordance with the performance is a practical approach to controlling the quality of the training dataset. 

\subsubsection{Inference and Iterative Refining.}
Table \ref{table_example} summarizes the iterative training process for fine-tuning the deep neural network using the S$^3$T method with the inspector's guidance. Refining the network for the 1\textsuperscript{st} iteration of the S$^3$T method was initiated with the last epoch of the previous iteration (i.e., the transfer learning) and continued for 20 more epochs. Then, the remainder of the dataset $V$, $V\setminus S_0$, is annotated by the refined network. Temporal coherence analysis is applied to $V\setminus S_0$, which contain objects with prediction scores between 0.5 and 0.9 to recover false negative results. This study considered 0.5 as the lower boundary of prediction threshold $t_l$ and 0.9 as the upper boundary $t_u$, which were found to minimize the volume of false negative results based on numerical experiments. The temporal coherence analysis in the 1\textsuperscript{st} iteration recovered 113 frames, and 37 frames of these were sampled according to SP(2), the sampling strategy considered by this study. In this study, $\alpha$ was 70\% in the 1$^\text{st}$ iteration, which means the inspector re-annotated 30\% of $S_1$ before adding them to the training dataset. In the 2$^\text{nd}$ iteration, the network was refined using the updated training dataset $T_2$ and then it was used to evaluate $V\setminus(S_0\cup S_1)$. Temporal coherence analysis recovered 79 images and 33 images were sampled and added to the training dataset. $\alpha$ was increased to 80\% of $S_2$ and the inspector annotated only 7 images out of the 33 before adding them to the training dataset. The iterative process was terminated after the 3\textsuperscript{rd} iteration of training when the target performance is achieved. 

\begin{table}[tb]
\small\sf\centering
\caption{Data sizes (\# images) in transfer learning (TL) and semi-supervised self-training (S$^3$T).}
\label{table_example}
\resizebox{\linewidth}{!}{
\begin{tabular}{l |r | rrr}
\toprule
& TL & \multicolumn{3}{c}{S$^3$T}\\
\cline{2-5}
\multicolumn{1}{c|}{$l$ : index of iterations} & 0  & 1 & 2 & 3 \\
\midrule
$T_l$: training dataset (Eq. 1) &  40 & 48 & 85 & 118 \\
$R_l$: recovered hard data samples & - & 113 & 79 & 50\\
$S_l$: a sampled subset of $R_l$       & - & 37 & 33 &  \\
$\alpha$: \% of $S_l$ for automatic annotation & - & 70\% & 80\% &\\
$S_{l,\alpha} $: automatically annotated data & - & 26 & 26 &  \\
$M_l$: manually annotated data & 8 & 11 & 7 &  \\
\bottomrule
\end{tabular}
}
\end{table}

\subsection{Quantitative Results}
In this study, experiments were conducted to evaluate the merits of the proposed method, the job efficiency of the developed assistive intelligence model, and its generalization capacity. 

\subsubsection{Object Detection Results.}
To evaluate the performance of object detection with the developed deep neural network, three standard evaluation matrices were used in this study:
\begin{itemize}
\item precision: it counts the number of correct predictions out of the total number of predictions;
\item recall: it counts the number of correct predictions out of total number of ground-truth objects;
\item f1-Score: it is the harmonic mean of precision and recall.
\end{itemize}

This study used the Intersection over Union (IoU) to determine whether a predicted object can be considered as a correct prediction. The IoU is the intersection between the predicted bounding box and the ground truth bounding box over the union of them. The ability of the network to correctly detect objects was evaluated on a range of IoU threshold values from 0.1 to 0.9 at a step of 0.1. The precision, recall, and f1-score from evaluating the test dataset $T_s$ are summarized in Table \ref{tab_result}. From the table, it can be observed that, given an IoU threshold value in the range of [0.1, 0.5], the Mask R-CNN that was initially transferred in for the bridge of inspection achieved the precision from 80.3\% to 87.5\%, the recall from 74.4\% to 81.0\%, and f1-score from 77.2\% to 84.1\%. The performance indicates the transferred network demonstrated some adaptability to the new task, but the amount of false negative detection is non-negligible. The performance of the network has not reached a satisfying level.

\begin{table}[tb]
\small\sf\centering
\caption{Performance (\%) of transfer learning (TL) and semi-supervised self-training (S$^3$T) in iterations.}\label{tab_result}
\centering
\begin{tabular}{l l| r | r  r  r  }
\toprule
&&\multicolumn{1}{c|}{TL}&\multicolumn{3}{c}{S$^3$T}\\
\cline{3-6}
\multicolumn{2}{r|}{Iteration, $l$} &0  & 1 & 2 & 3\\
\midrule
\multicolumn{1}{c}{IoU}&&&&&\\
\multirow{3}{*}{0.1} & Precision & 87.5  & 86.3 & 94.0 & 93.45  \\
& Recall & 81.0 & 95.4 & 93.5 & 95.30 \\
& F1-Score & 84.1  & 90.6 & 93.8 & 94.4\\
\hline
\multirow{3}{*}{0.2} & Precision & 87.1  & 85.7 & 93.9 & 93.5 \\
& Recall & 80.7 & 94.8 & 93.3 & 95.3\\
& F1-Score & 83.4 & 90.0 & 93.6 & 94.4 \\
\hline
\multirow{3}{*}{0.3} & Precision & 86.8  & 85.2 & 93.7 & 93.4  \\
& Recall & 80.3 & 94.2 & 93.2 & 95.2\\
& F1-Score & 83.4 & 89.5 & 93.4 & 94.3 \\
\hline
\multirow{3}{*}{0.4} & Precision & 84.6 & 84.2 & 93.2 & 93.1  \\
& Recall & 78.3 & 93.1 & 92.7 & 94.9\\
& F1-Score & 81.3 & 88.5 & 93.0 & 94.0 \\
\hline
\multirow{3}{*}{\textbf{0.5}} & \textbf{Precision} & 80.3 & 81.7 & 90.7 & 91.8\\
& \textbf{Recall} & 74.4  & 90.3 & 90.1 & 93.6\\
& \textbf{F1-Score} & 77.2  & 85.8 & 90.4 & 92.7  \\
\hline
\multirow{3}{*}{0.6} & Precision & 75.9 & 77.4 & 85.7 & 88.5\\
& Recall & 70.2 & 85.6 & 85.1 & 90.2\\
& F1-Score & 73.0 & 81.3 & 85.4 & 89.3 \\
\hline
\multirow{3}{*}{0.7} & Precision & 65.4 & 66.6 & 74.6 & 78.1   \\
& Recall & 60.5 & 73.6 & 74.2 & 79.6 \\
& F1-Score & 60.5 & 73.6 & 74.2 & 79.6 \\
\hline
\multirow{3}{*}{0.8} & Precision & 43.7 & 43.8 & 50.1 & 49.0  \\
& Recall & 40.5 & 48.5 & 49.8 & 49.9  \\
& F1-Score & 42.1 & 46.0 & 49.9 & 49.5 \\
\hline
\multirow{3}{*}{0.9} & Precision & 6.3 & 3.0 & 6.7 & 4.6  \\
& Recall & 5.9 & 3.3 & 6.7 & 4.7 \\
& F1-Score  & 6.1 & 3.1 & 6.7 & 4.7  \\
\bottomrule
\end{tabular}
\end{table}

Therefore, the network was iteratively refined using the proposed S$^3$T method with human-in-the-loop to seek further improvement. After being re-trained in the 1st iteration, the recall was effectively increased by 15\%, approximately. For example, when the IoU threshold value is 0.5, the precision increases from 80.3\% to 81.7\%, and recall becomes 90.3\% from 74.4\%, yielding a 85.8\% f1-score after the 1st iteration. The changes indicate that $M_0$, the additional small set of manually annotated hard samples added to the training dataset, effectively improves the ability to correctly detect more objects. The performance of the network has met the requirement after being refined for additional two iterations, reaching 91.8\% precision, 93.6\% recall, and 92.7\% f1-score at the IoU threshold value 0.5. The S$^3$T method has effectively brought the performance of the network to a satisfying level. As the IoU threshold value decreases gradually from 0.5 to 0.1 at a step size of 0.1, the evaluation becomes less conservative. Consequently, fewer false negative detections are rendered by the network but maybe more false positive detections. On the other hand, selecting a higher IoU threshold value makes the evaluation more conservative. As it increases gradually from 0.5 to 0.9 at a step size of 0.1, the f1-score is diminishing, signifying the reduction of both precision and recall values. In this application setting, false positive detections are less concerned than false negative detections. This is because the inspector will retrieve and analyze frames that contain detected and segmented objects that s/he wants to inspect. Therefore, false positive detections can be found and eliminated by the inspector. But, false negatives are more critical because inspectors cannot overlook any potential damages. From the analysis above, it can be inferred that this S$^3$T method is very applicable to the development of the proposed assistive intelligence model for detecting bridge elements from inspection videos.

Another important observation from the table is the relationship between the IoU threshold value and the recall value during the iterative process of network fine-tuning. A variation of a recall value within a range of 6.6\% has been observed after the initial adaption through transfer learning when the IoU threshold value increases from 0.1 to 0.5. However, this variation reduces in each successive iteration. For example, after the third iteration, this variation reduces to 1.6\%. Precision and recall values at any of the IoU threshold values increase over iterations and reach the maximum after the 3rd iteration. For example, at the IoU threshold value 0.5, f1 score increases about 8.6\%, 4.6\%, and 2.3\%, respectively, from their previous iteration. This means the network learns new features from each iteration and gradually moves toward the learning limit. The improvement rate is diminishing during the iterative process. When the IoU threshold value continues increasing from 0.5 and onward, the recall value in any iteration drops rapidly and has been less than 10\% when the IoU threshold value is 0.9. Moreover, the increasing trend of the recall value over iterations slows down quickly at the IoU threshold value 0.6. When the IoU threshold value is greater than 0.6, the increasing trend of recall over iterations is rapidly flatted out and becomes a decreasing trend. The reason for this sharp decrease of recall value along with the increase of the IoU threshold value is that the network considers a detected object as a true positive detection only if the overlap between the ground truth and the bounding box of the detected object is very high, which increases the amount of false negative detections and decreases the amount of true positive detections.

\subsubsection{Instance Segmentation Results.} The study also evaluated the quality of the proposed assistive intelligence model in segmenting bridge elements from inspection videos. The mask Intersection over Union (IoU) is a measure of segmentation quality, which is the ratio of the overlap between the predicted segmentation mask and the ground truth mask to the union of these two masks. The predicted segmentation mask is considered as a true positive prediction if the mask IoU value is no less than a pre-specified threshold. The higher the threshold value is selected for the evaluation, the stricter the evaluation becomes. Accordingly, The precision for each class can be calculated. After that, the average of the class-level precision values, named mean precision and denoted as mP, is determined.

\begin{figure}[htbp]
\centering
\includegraphics[width=\linewidth]{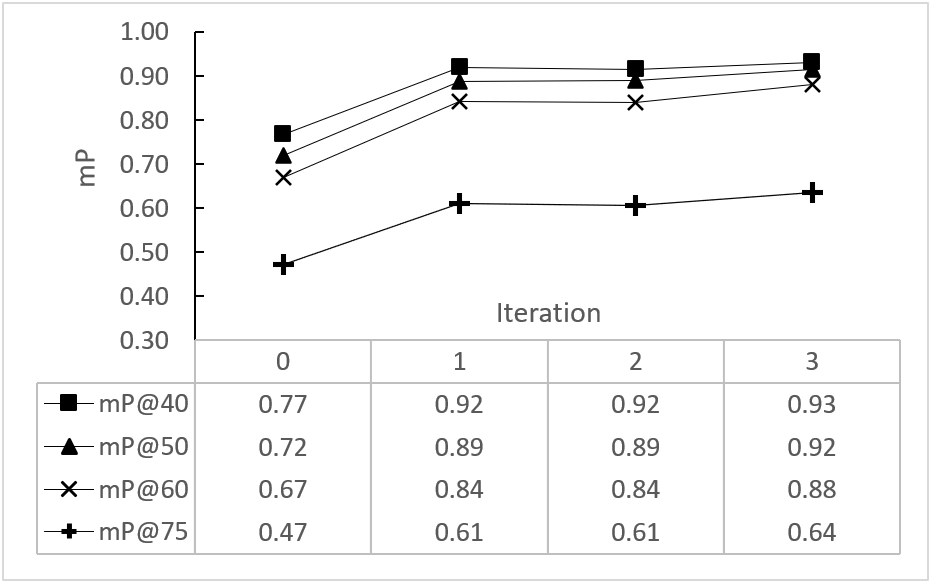}
\caption{The mean Precision (mP) over iterations at different mask IoU threshold values.}
\label{fig_map}
\end{figure}

Figure \ref{fig_map} shows the curve of mP value during the iterative process for fine-tuning the network at four levels of mask IoU threshold value, wherein the x-axis represents the number of iterations and the y-axis represents the mP value. The plot shows the mP curve at the mask IoU threshold value 0.4 is an upwarding curve on the top of other curves. The mP value at the end of the iterative process reaches 93\%. When the mask IoU threshold value increases to 0.5, the curve just drops slightly and the shape of the curve has no change. The mP value at the end of the iterative process reaches 92\%. However, with a larger mask IoU threshold value such as 0.75, the mP curve clearly drops to a lower position. This is because the amount of true positive results at a larger mask IoU threshold value is low although the total number of correctly segmented objects increases over iterations. 

\subsubsection{Efficiency of Transfer Learning.}

To demonstrate the high cost-effectiveness of transfer learning for the initial adaption, this study trained a Mask R-CNN from the scratch using 144 annotated images. Results of the comparison are summarized in Table \ref{tab:transfer_learning}. The first 600 epochs for training the network from scratch took 13.2 hours, and the network performance (32.3\% precision, 18.3\% recall, and 23.4\% f1 with an IoU threshold value 0.5) is well below the target performance when it was tested on the dataset $T_s$. The experiment clearly demonstrates the network requires a huge number of training samples to be trained from the scratch, which is infeasible for developing the desired model of inspection video data analysis due to the scarcity of labeled data. The proposed transfer learning used only 40 annotated images as the training dataset and took only 20 minutes to transfer the capability of an existing Mask R-CNN in multiclass object detection and segmentation to the new task with bridge elements. The performance of the transferred Mask R-CNN has a much better result (80.3\% precision, 74.4\% recall, and 77.2\% f1 at the IoU 0.5). The comparison summarized in Table \ref{tab:transfer_learning} demonstrates that transfer learning reduced the training time by at least 95\% and has improved the performance of the network tremendously. 

\begin{table}[bt]
\small\sf\centering
    \caption{Cost-effectiveness of transfer learning in comparison with training from the scratch.}
    \label{tab:transfer_learning}
\resizebox{\linewidth}{!}{
    \begin{tabular}{p{2.8cm}|>{\centering\arraybackslash}p{1.3cm}|>{\centering\arraybackslash}p{1cm}>{\centering\arraybackslash}p{0.7cm}>{\centering\arraybackslash}p{0.4cm}}
    \toprule
    & Training time (hrs) & Precision (\%) & Recall (\%)& F1 (\%) \\
    \midrule
    Training from scratch & 13.2  & 32.3 & 18.3 & 23.4 \\
    \hline
    Transfer learning & 0.33  & 80.3 & 74.4 & 77.2 \\
    \bottomrule
    \end{tabular}
   }
\end{table}

\subsubsection{Comparison with a State-of-the-Art Method.}

\begin{table*}[tb]
\small\sf\centering
    \caption{Comparison of the proposed approach to directly transferring a Mask R-CNN with various volumes of training dataset.}
    \label{tab:comparison_study}
    \begin{tabular}{l|>{\raggedleft}p{2.4cm}|>{\raggedleft}p{1.5cm}>{\raggedleft}p{1.4cm}|>{\centering\arraybackslash}p{2.3cm}|>{\centering\arraybackslash}p{1.2cm}>{\centering\arraybackslash}p{1cm}>{\centering\arraybackslash}p{1.5cm}}
\toprule
Method & No. of manually annotated images& Annotation time (min)& Training time (min) & Inference speed (sec/frame) & Precision (\%) & Recall (\%) & F1-score (\%)\\
\midrule
Mask R-CNN & 22& 72 & 18 & 0.55  & 68.0 & 68.4 & 68.2\\
Mask R-CNN & 44 & 143 & 20 & 0.55  & 82.0 & 79.0 & 80.5\\
Mask R-CNN & 220 & 715 & 33 & 0.55  & 85.8& 91.8& 88.7\\
Mask R-CNN & 440 & 1,430 & 66 & 0.55  & 89.7& 92.3& 91.0\\
Self-training & 48  & 156  & 72  & 0.55  & 88.9  & 76.7 & 82.4 \\
\midrule
Our approach & \bf{66} & \bf{215} & \bf{72} & \bf{0.55} & \bf{91.8} & \bf{93.6} & \bf{92.7} \\
\bottomrule
\end{tabular}
\end{table*}
This study compared the proposed approach (i.e., transfer learning plus S$^3$T with human-in-the-loop) to the Mask R-CNN adapted with transfer learning only and the traditional self-learning, from the perspectives of annotation time, training time, inference speed, and accuracy. To show the reliance of the performance of the transferred Mask R-CNN on the volume of training dataset, this study measured the performance of the Mask R-CNN after independently transferred with four random training datasets: 0.5\% (22 images) , 1\% (44 images), 5\% (220 images) and 10\% (440 images) of images in Dataset D. Results from the comparison are summarized in Table \ref{tab:comparison_study}. It is observed that the transfer learning by itself can improve the performance of the network, but the improvement is at a rapidly increasing cost of annotation time. Transferring the Mask R-CNN with 440 annotated images took 1430 minutes (i.e., 23.8 hours) for data annotation and 66 minutes for training. This network achieves 89.7\% precision, 92.3\% recall, and 91.0\% f1 score, close to the performance of the proposed approach in this paper. The proposed approach reduces the annotation time by 85\% with a comparable training time (only 6 minutes longer), and it achieves a better performance (91.8\%  precision, 93.6\%  recall, and 92.7\% f1 score). Self-training does not use the inspector guidance in the iterative re-training process, thus saving about one hour of annotation time compared to the S$^3$T with human-in-the-loop. But the performance is not satisfying. The inference speed of all the models is 0.55 seconds per frame. This comparative study demonstrated that the $S^3T$ method with human-in-the-loop is more cost-effective compared to directly transferring the Mask R-CNN. It also significantly improves the model performance compared to traditional self-training, due to the engagement of experienced inspectors in the iterative re-training process. The impact of human-in-the-loop on self-training is further examined.

\subsubsection{The Impact of Human-in-the-Loop on Self-Training.}
\label{subsubsec:alpha}

The S$^3$T method that keeps human-in-the-loop is a combination of self-training and active learning. The portion of additional training data re-annotated by experienced inspectors in each iteration of the self-training process may impact the efficiency of model development and the performance of the resulting final model. This study used four experiments to illustrate the impact of human-in-the-loop on self-training, which are summarized in Table \ref{tab:alpha}. The four experiments all began with the same initial model whose performance is 80.3\% precision, 74.4\% recall, and 77.2\% f1-score. In all the experiments the model is trained for three iterations. Inspectors annotated eight images to re-train the initial model for the first iteration. With this iteration, the performance is increased to 81.7\% precision, 90.3\% recall, and 85.8\% f1-score. After that, the four experiments differ in the ratio of model-annotated additional training data ($\alpha$). If an experiment yields satisfying model performance (i.e., both recall and precision are at least 90\%, and f1-score is at least 92\% at the IoU threshold value 0.5) within three iterations, the final performance is highlighted as bold.
\begin{table}[tb]
\small\sf\centering
    \caption{Variants of inspector engagement in self-training }
    \label{tab:alpha}
    \begin{tabular}{c|l|rrrr}
    \toprule
    &\multicolumn{1}{|c|}{iterations}&0&1&2&3\\
    \midrule
    \multirow{6}{*}{1}&precision (\%)&80.3 &81.7 &90.1 &\bf{91.8} \\
    &recall (\%)&74.4&90.3&90.7 &\bf{93.6} \\
    &f1-score (\%)&77.2 &85.8 &90.4 &\bf{92.7} \\
    &$S_{l,\alpha}$ (frame)&0&26&26&\\
    &$M_l$ (frame)&8&11&7&\\
    &$\alpha_l$ (\%)&0 &70 &80 & \\
    \midrule
    \multirow{6}{*}{2}&precision (\%)&80.3 &81.7 &91.8&\bf{94.5} \\
    &recall (\%)&74.4&90.3&84.7&\bf{91.5} \\
    &f1-score (\%)&77.2 &85.8 &88.1&\bf{93.0} \\
    &$S_{l,\alpha}$ (frame)&0&22&26&\\
    &$M_l$ (frame)&8&15&11&\\
    &$\alpha_l$ (\%)&0 &60 &70 & \\
    \midrule
    \multirow{6}{*}{3}&precision (\%)&80.3 &81.7 & 90.1& 91.7 \\
    &recall (\%)&74.4&90.3& 90.7& 92.1 \\
    &f1-score (\%)&77.2 &85.8 & 90.4& 91.9 \\
    &$S_{l,\alpha}$ (frame)&0&26&28&\\
    &$M_l$ (frame)&8&11&3&\\
    &$\alpha_l$ (\%)&0 &70 &90 & \\
    \midrule
    \multirow{6}{*}{4}&precision (\%)&80.3 &81.7 & 88.3 &88.9 \\
    &recall (\%)&74.4&90.3& 75.7& 76.7\\
    &f1-score (\%)&77.2 &85.8 &81.5 &82.4 \\
    &$S_{l,\alpha}$ (frame)&0&37&33&\\
    &$M_l$ (frame)&8&0&0&\\
    &$\alpha_l$ (\%)&0 &100 &100 & \\
    \bottomrule
    \end{tabular}
\end{table}

Experiment 1 is our approach in Tables \ref{tab_result} and \ref{tab:comparison_study}. Compared to experiment 1, experiment 2 let the inspector re-label more data for the second and third iterations of training. It not only provided a satisfying performance at the end of iteration 3 but increased the f1-score with a small margin over that from experiment 1. Compared to experiment 1, experiment 3 used less inspector-annotated additional training data for the third iteration. Although it does not achieve the target performance at the end of the third iteration, f1-score is just below the target by 0.1\% a tiny margin. That is, the target performance is likely to be reached in the next iteration of training. Unlike experiments 1$\sim$3 that are all self-training with human-in-the-loop, experiment 4 is the traditional self-training without human-in-the-loop. That is, the model teaches itself by adding additional data to the training dataset iteratively. The added new data are those with the most confident label prediction by the model. To make it a fair comparison, experiment 4 used the same amount of additional training data as experiment 1 in each iteration except that the additional data for iterations 2 and 3 are all model-annotated. By the end of the third iteration, the model performance is well below the target performance. The drop of the model performance from the first iteration to the second iteration, and the slow increase of the performance from the second iteration to the third iteration, indicate including the inspector’s guidance in self-training is critical.

The proposed S$^3$T method with human-in-the-loop (in experiments 1$\sim$3) has tremendously improved the efficiency of model development as it is compared to the supervised learning method (Mask R-CNNs in Table \ref{tab:comparison_study}). Its performance improvement has a large margin over the traditional self-training (in experiment 4). Further optimizing the engagement of inspectors in self-training has just incremental improvement against the heuristic strategy of this paper.

\subsubsection{Job Efficiency of the Assistive Intelligence Model.}

A full image usually contains multiple elements of the bridge. An inspector needs to search and find all the elements, segment each identified element by marking its boundary, and provide the object name. On average, each image in this study contains 15 objects of different sizes and shapes. To provide better quality, a polygon rather than a rectangle is preferred for segmenting identified bridge elements. Drawing a tight polygon on a single object may require defining 15$\sim$30 points on the image. On average, it took around 3.25 minutes to detect and manually segment bridge elements in a full image in this study. Not to mention that issues related to human factors, such as the fatigue developed from repeatedly working on high cognitive tasks, further lengthen the time required for manually analyzing the big video data collected from bridge inspection. The approach that this paper proposes requires inspectors to analyze a small amount of data manually, helping the network achieve satisfying performance in object detection and segmentation. The automated data processing relieves inspectors from the time-consuming labor work. The developed assistive intelligence model can finish the same job with very high accuracy but with only 0.55 seconds per image, which is 350 times faster than the manual approach. The impact of the improved work efficiency is tremendous because a real-world task usually requires analyzing hundreds of thousands of images.

In the real-world implementation, the developed assistive intelligence model will detect and segment bridge elements from the inspection video data. Then, the inspector retrieves the elements of interests from the large pool of video frames and evaluates damages or other defects associated with them. The developed model assists inspectors in that, it reduces the human effort in searching and finding the needed data so that inspectors can focus on knowledge-intensive tasks. Moreover, by removing the burden of browsing hours of videos to look for bridge elements for evaluation, inspectors are less likely to work in a state of fatigue or lacking focus. 

\subsubsection{Generalization Capability of the Proposed Method.}

To assess if the developed network is applicable to other bridges, the study used it to detect and segment the same ten classes of bridge elements of another two bridges, named bridges A and B. Correspondingly, the bridge that has had a network developed for is named bridge C. When the network built for bridge C was applied to bridges A and B, the performance of it is comparable to that of the initial network for bridge C. The trained network achieves 76.8\% precision, 73.0\% recall, and 74.8\%  f1-score for bridge A; and for bridge B it accomplishes 61.2\% precision, 60.0\% recall, and 60.6\% f1-score. Bridge A is more similar to bridge C than bridge B. Therefore, the network trained for bridge C performs better in analyzing the inspection data of bridge A than bridge B. 
Therefore, the network for bridge C has a certain degree of generalization, and it is good enough to serve as the initial network for other bridges. Further implementing the S$^3$T method developed in this paper will adapt the assistive intelligence model developed for bridge C to bridges A and B, respectively, to achieve the target performance on these bridges.

\subsection{Qualitative Results}

The developed assistive intelligence model was tested on the inspection dataset $D$. Some qualitative examples selected from the testing result are illustrated below.

\subsubsection{An Illustrative Example of Temporal Coherence Analysis.}
\begin{figure*}[tb]
\centering
\subfloat{\includegraphics[width=6in]{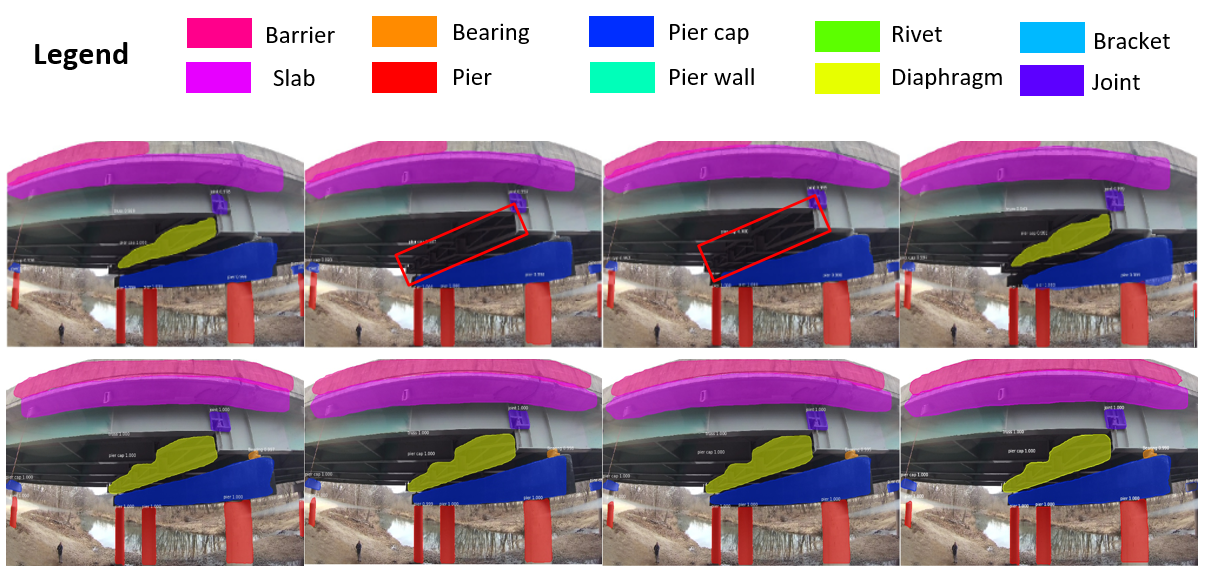}}
\hfil
\caption{An example of temporal coherence analysis.}
\label{fig_comparison}
\end{figure*}

Figure \ref{fig_comparison} illustrates an example wherein the temporal coherence analysis improves the model performance by eliminating false negative results. In the first row of Figure \ref{fig_comparison}, the single-image based network correctly predicted the diaphragm in the 1\textsuperscript{st} and the 4\textsuperscript{th} frames, however, failed to detect it in the 2\textsuperscript{nd} and the 3\textsuperscript{rd} frames. The second row is the result after applying the temporal coherence analysis, which shows that the diaphragm was correctly segmented in all of the four frames. Note that false negative results are more severe than false positive results in the task of bridge inspection. Because false positives rendered by the deep network can be re-checked by the inspectors but false negatives ignored by the network will not have such an opportunity. Therefore, effectively reducing false negative results is particularly more important for the bridge inspection. 

\subsubsection{Representative Examples of Successful Detection and Segmentation.}
Figure \ref{fig_visual-result} (a) illustrates some representative examples of successful segmentation of bridge elements by the developed network. The first column of Figure \ref{fig_visual-result} (a) is an example that a partial joint of different scales in three images is detected and segmented correctly from all images. The second column illustrates the capability of detecting and segmenting a rivet from various views. The network successfully detects and segments the rivet in a low light condition, as the second figure in column two illustrates. The third column comprises successful examples of segmenting the rivet at a wide range of scale variations. The fourth column shows that the developed network is successful in detecting and segmenting multiple objects at various distances in complex scenes.

\subsubsection{Representative Examples of False Negative Detection}
\begin{figure*}[bt]
\centering
\subfloat{\includegraphics[width=\linewidth]{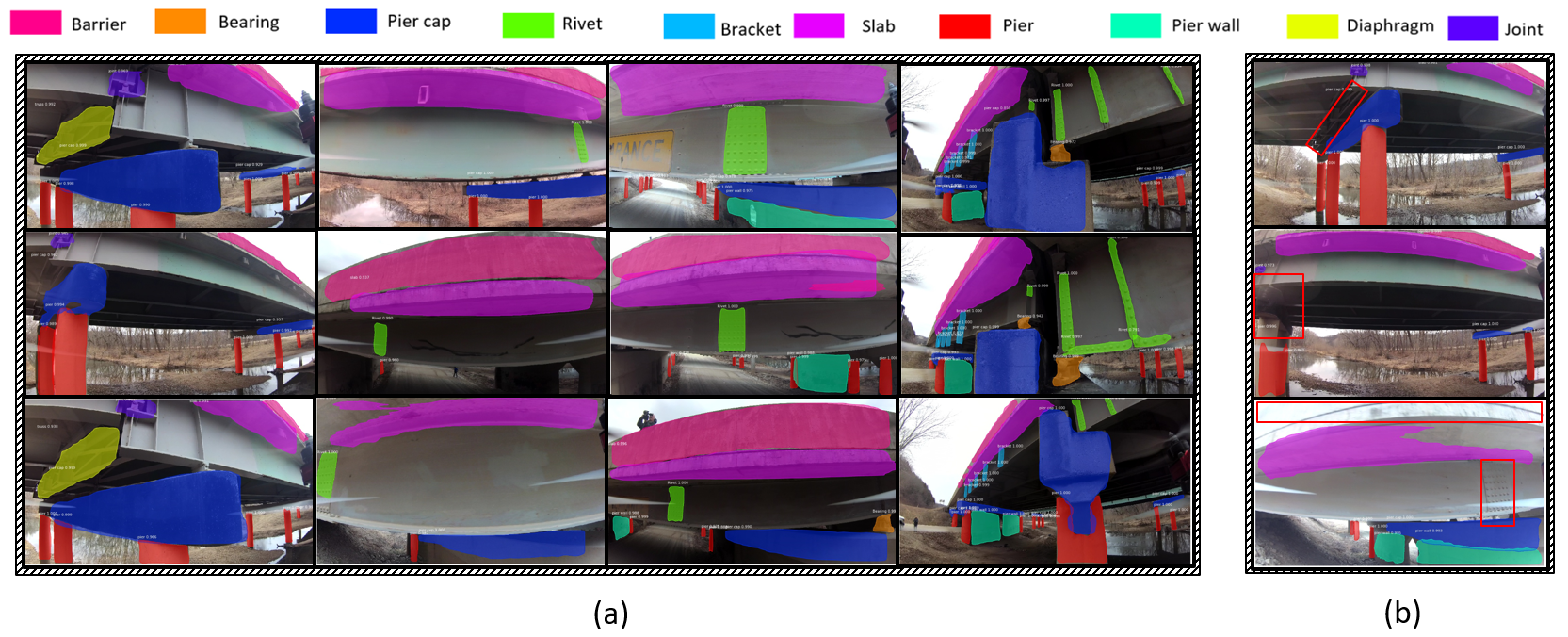}}
\hfil
\caption{Examples of (a) successful detection and segmentation, and (b) false negative detection.}
\label{fig_visual-result}
\end{figure*}

The red bounding boxes in Figure \ref{fig_visual-result} (b) represent false negative results by the developed network. The diaphragm in the top frame and the pier cap in the middle frame are difficult to recognize because of the dark illumination level. In the bottom frame, the developed network fails to detect the barrier due to the high exposure. The rivet in the bottom frame is not detected because of its similar appearance with the background. External illumination sources on inspection platforms and image contrast enhancement techniques are potential solutions to overcome these challenges.

\section{Conclusions}
\label{sec:Conclusion}
This paper presents a method to develop an assistive intelligence model to support bridge inspectors in segmenting multiclass bridge elements from big complex video data collected by aerial inspection platforms. With a small initial training dataset annotated by inspectors, a Mask R-CNN pre-trained on a large public dataset was transferred to the new task of bridge inspection. Then, the temporal coherence analysis was used to recover false negative results and thus identify the weakness of the current network to improve, which adds a nearly negligible additional computation load during the inference compared to other methods based on the motion guidance. A semi-supervised self-training algorithm was developed to engage the inspector in refining the network iteratively. The domain knowledge of the inspector quickly brought the network's performance to a satisfying level.

Assessment results of the developed assistive intelligence model showed that the proposed approach to the model development uses a small amount of time and guidance from bridge inspectors to achieve a high performance in segmenting multiclass structural elements from the big complex inspection videos. For example, the developed model has achieved around 94\% of precision, 92\% of recall, and 92\% mAP when the IoU threshold value is 0.5. The study revealed that having sufficient guidance from experienced bridge inspectors, particularly in early iterations of the semi-supervised self-training for refining the network is critical for maintaining the quality of the training dataset. The amount of human annotation can be gradually reduced as the network becomes more reliable in performing its tasks. 

The paper has identified rooms for improvement. Adapting the assistive intelligence model to bridges with additional structural elements is the next step to extend this paper. One important future work is to improve the inference speed. While the developed model is able to achieve a high performance with a small amount of human hours and the computation time for training the network, improving the testing speed to have the real time inference capability is highly desired. Moreover, contextual information and the spatial correlation among objects could be utilized to further improve the segmentation accuracy. Another Future work will evaluate the change in cognitive load and other psychological states of inspectors assisted by the assistive intelligence in bridge inspection. The evaluation involves using biometric sensors such as eye movement trackers and electroencephalogram (EEG) to detect heavy cognitive load, fatigue, and lost of focus of inspectors in their tasks.

\begin{acks}
Research presented in this paper is partially supported by the U.S. Department of Transportation, Office of the Assistant Secretary for Research and Technology (USDOT/OST-R) under Grant No. 69A3551747126 through INSPIRE University Transportation Center (\url{http://inspire-utc.mst.edu}). The views, opinions, findings and conclusions reflected in this publication are solely those of the authors and do not represent the official policy or position of the USDOT/OST-R, or any State or other entity.
\end{acks}

\section*{Appendix}
\label{sec_appendix}

\begin{figure}[htbp]
\centering
\includegraphics[width=\linewidth]{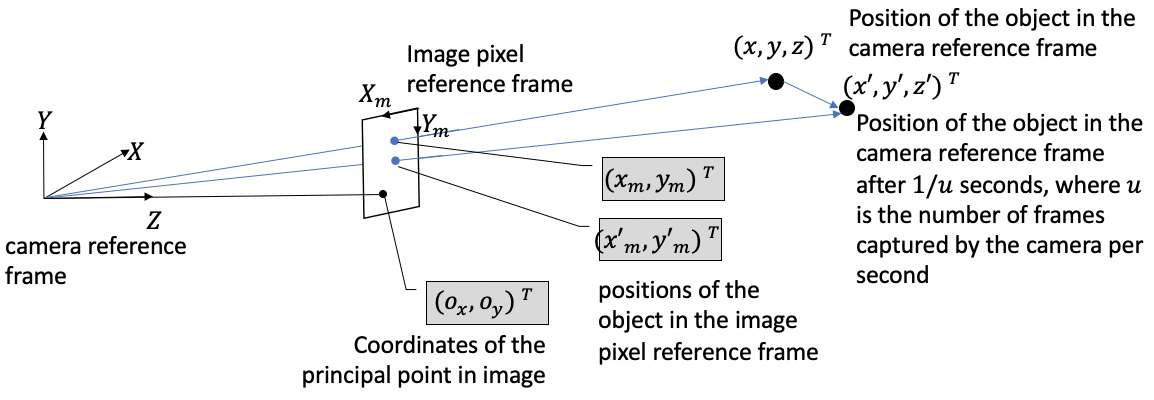}
\hfil
\caption{Intrinsic camera parameters define the relationship between the camera reference frame and the image pixel reference frame.} 
\label{fig_appendix}
\end{figure}

Figure \ref{fig_appendix} below describes an object in the 3D camera reference frame and the 2D image pixel reference frame. At a certain time, the coordinates of the object in the camera reference frame is $(x,y,z)^T$, and $(x_m,y_m)^T$ in the image pixel reference frame. The relationship of these two sets of coordinates of the object is determined as:
\begin{equation}
    \left[
    \begin{array}{l}
         x_m  \\
         y_m 
    \end{array}
    \right]
    =
    \left[
    \begin{array}{l}
         o_x  \\
         o_y 
    \end{array}
    \right]
    -\frac{f}{s_m}
    \left[
    \begin{array}{l}
         x/z  \\
         y/z 
    \end{array}
    \right]
\end{equation}
where $(o_x,o_y)^T$ represent the coordinates of the principal point (in pixel) of images, $f$ is the focal length of camera, and $s_m$ is the size of pixel (millimeter per pixel and it is assumed to be the same on both $X_m$ and $Y_m$ axes in this study). $(o_x,o_y)^T$, $f$, and $s_m$ are intrinsic camera parameters. 

In this study, the object is static, and the camera is moving, in the world reference frame. But in the camera reference frame, the object is moving. Imaging that the object moves from $(x,y,z)^T$ to $(x',y',z')^T$ in $1/u$ seconds in the camera reference frame, where $u$ is the speed of the camera in capturing images (i.e., how many images are taken per second). Accordingly, the object moves from $(x_m,y_m)^T$ to $(x'_m,y'_m)^T$ in the next frame of image. $(x',y',z')^T$ must be within an L2 ball of radius $\epsilon$ centered at $(x,y,z)^T$. The radius of the ball is determined by the maximum linear and rotational speeds of the camera. Therefore, the displacement of the object from one frame of image to the next frame is:
\begin{equation}
    \begin{aligned}
        \Delta r &= \sqrt{(x'_m-x_m)^2+(y'_m-y_m)^2}\\
        &=\frac{f}{s_m}\sqrt{\left(\frac{x'}{z'}-\frac{x}{z}\right)^2+\left(\frac{y'}{z'}-\frac{y}{z}\right)^2}\\
        &\leq \frac{f}{s_m}\frac{\epsilon}{\min(|z|,|z'|)}=\Delta d
    \end{aligned}
    \label{eq:Deltad}
\end{equation}
Eq. \ref{eq:Deltad} shows the maximum displacement of an object ($\Delta d$) from one frame to the next frame of image is proportional to the focal length ($f$) and the radius of the L2 ball that defines the boundary of the object’s relative motion rate in the camera reference frame ($\epsilon$). $\Delta d$ is inversely proportional to the size of pixel $s_m$ and the minimum distance of the camera to the object along the optical axis when the two successive frames are captured ($\min(|z|,|z'|)$). Eq. \ref{eq:Deltad} indicates $\Delta d$ is affected by intrinsic camera parameters $f$ and $s_m$. Extrinsic parameters affect $\Delta d$ too because the term $\epsilon/\min(|z|,|z'|)$ is determined by the position and motion rate of the camera in the world reference frame.

\bibliographystyle{SageV}
\bibliography{ref.bib}

\end{document}